\documentclass[letterpaper, 10 pt, conference]{ieeeconf}  

\IEEEoverridecommandlockouts                              

\usepackage{epsfig} 
\usepackage{times} 
\usepackage{amsmath} 
\usepackage{amssymb}  
\usepackage{comment}
\usepackage{graphicx}
\usepackage{booktabs}
\usepackage[percent]{overpic}
\usepackage{dcolumn}
\usepackage{url}
\usepackage[T1]{fontenc}

\newcommand{\pv}{\mathbf{p}}
\newcommand{\av}{\mathbf{a}}
\newcommand{\gv}{\mathbf{g}}
\newcommand{\bv}{\mathbf{b}}
\newcommand{\Am}{\mathbf{A}}
\newcommand{\Rm}{\mathbf{R}}
\newcommand{\Hm}{\mathbf{H}}
\newcommand{\Um}{\mathbf{U}}
\newcommand{\Vm}{\mathbf{V}}
\newcommand{\Pm}{\mathbf{P}}
\newcommand{\Qm}{\mathbf{Q}}
\newcommand{\omegav}{\boldsymbol{\omega}}
\newcommand{\Lambdam}{\boldsymbol{\Lambda}}
\DeclareMathOperator*{\argmin}{arg\,min}

\title{\LARGE \bf
Inertial-Based Scale Estimation for Structure from Motion\\ on Mobile Devices
}

\author{Janne Mustaniemi$^{1}$, Juho Kannala$^{2}$, Simo S\"arkk\"a$^{2}$, Jiri Matas$^{3}$ and Janne Heikkil\"a$^{1}$
\thanks{$^{1}$Center for Machine Vision and Signal Analysis, University of Oulu, Finland
        {\tt\small janne.mustaniemi@oulu.fi}}%
\thanks{$^{2}$Aalto University, Finland }%
\thanks{$^{3}$Centre for Machine Perception, Department of Cybernetics, Czech Technical University, Czech Republic}%
}
\begin{document}

\maketitle
\thispagestyle{empty}
\pagestyle{empty}

\begin{abstract}

Structure from motion algorithms have an inherent limitation that the reconstruction can only be determined up to the unknown scale factor. Modern mobile devices are equipped with an inertial measurement unit (IMU), which can be used for estimating the scale of the reconstruction. We propose a method that recovers the metric scale given inertial measurements and camera poses. In the process, we also perform a temporal and spatial alignment of the camera and the IMU. Therefore, our solution can be easily combined with any existing visual reconstruction software. The method can cope with noisy camera pose estimates, typically caused by motion blur or rolling shutter artifacts, via utilizing a Rauch-Tung-Striebel (RTS) smoother. Furthermore, the scale estimation is performed in the frequency domain, which provides more robustness to inaccurate sensor time stamps and noisy IMU samples than the previously used time domain representation. In contrast to previous methods, our approach has no parameters that need to be tuned for achieving a good performance. In the experiments, we show that the algorithm outperforms the state-of-the-art in both accuracy and convergence speed of the scale estimate. The accuracy of the scale is around $1\%$ from the ground truth depending on the recording. We also demonstrate that our method can improve the scale accuracy of the Project Tango's build-in motion tracking.

\end{abstract}

\section{INTRODUCTION}
Structure from motion (SfM) is the process of estimating 3D structure and camera motion from a series of 2D images. The scale ambiguity is a well-known limitation of this process. The reconstruction is only possible up to an unknown scale factor when using a monocular camera. However, the scale information would often be useful, for example, when making body size measurements for online shopping. Similarly, the scale information could be utilized in 3D printing. The user could first scan the object with a smart device and then print the object in exact dimensions with a 3D printer.

The scale ambiguity can be solved by using at least two calibrated cameras or a depth camera. Besides the fact that the stereo and depth cameras have a limited operational range, they are also more expensive and rarely included in mobile devices. The global positioning system (GPS) can also be used for obtaining the metric scale of the reconstruction. However, the GPS is typically relatively inaccurate and only works outdoors.

Some scale estimation methods avoid the need for extra hardware by making assumptions about the scene content. For instance, the smart device application \cite{rulerphone} allows the user to make metric measurements from the scene by using the known dimensions of a credit card, which is embedded to the scene. Similarly, the application \cite{thirdlove} detects the device itself from the mirror in order to make body size measurements of the user. The method \cite{smartmeasure} solves the problem by assuming that the ground is flat and that the approximate height of the camera from the ground is known.

\begin{figure}[t]
 \includegraphics[width=0.48\textwidth]{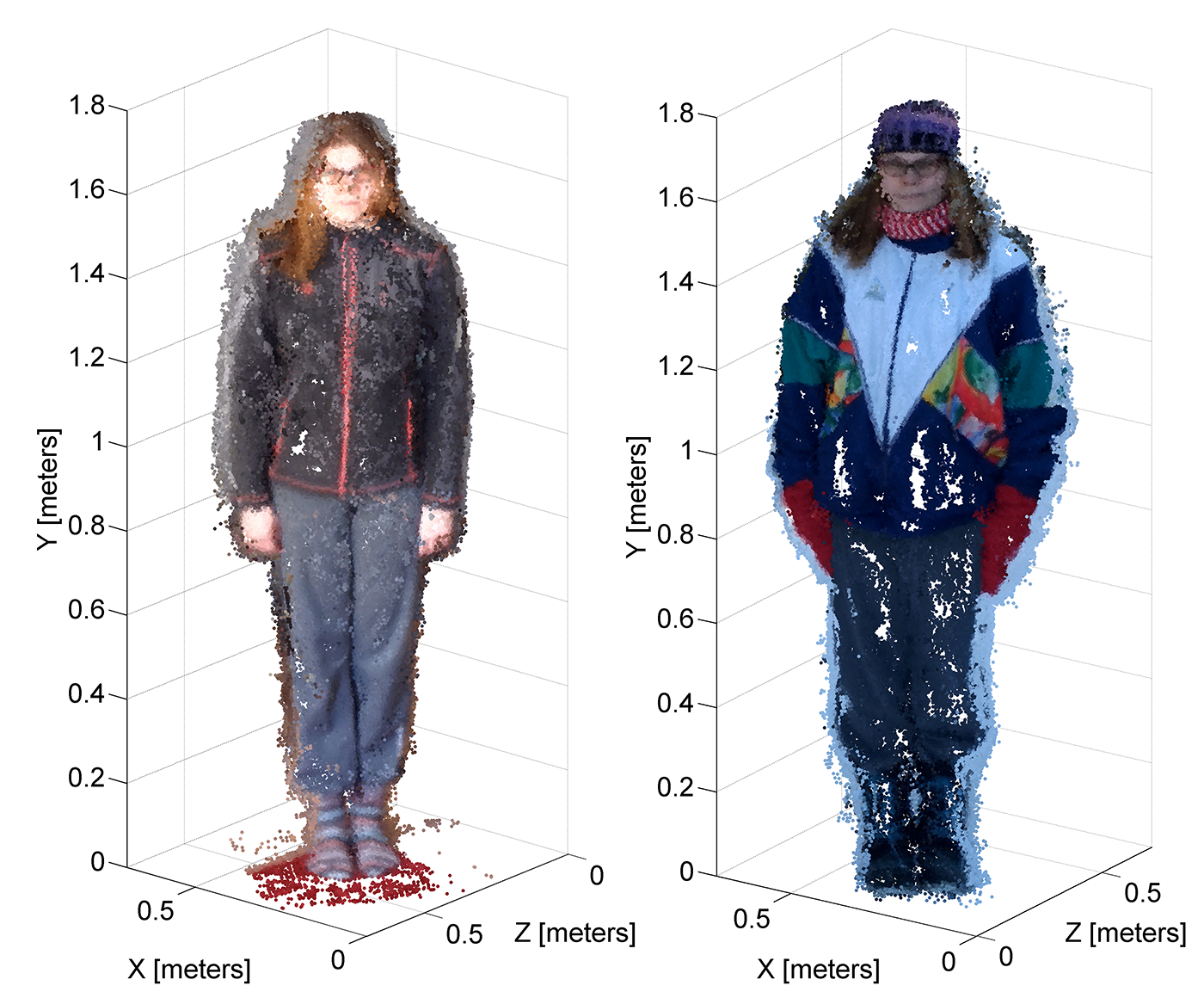}
\caption{Visual reconstructions after the scale correction. The reconstructions have been rotated so that the gravity vector, which is another output of our method is aligned with the y-axis.}
\label{fig:1}
\end{figure}

In this paper, we propose a method for recovering the metric scale of a visual SfM reconstruction by using inertial measurements recorded with an IMU that is rigidly attached to the camera. An example of the scaled reconstruction is shown in Figure \ref{fig:1}. The proposed approach outperforms the state-of-the-art, especially when dealing with noisy measurements, e.g.\ due to motion blur, rolling shutter artifacts or low-quality IMU. The accuracy of scale is typically around $1\%$ from the ground truth depending on the recording. We also present a calibration method, which aligns the inertial and visual measurements both temporally and spatially. Therefore, the algorithm can be easily bundled with any structure from motion software that outputs the camera poses. Our method and data will be made available as open source upon publication of the paper.

\section{RELATED WORK}
The fusion of visual and inertial measurements has been a popular research topic in the robotics community. Most previous systems are focused on real-time tracking and navigation, e.g.\ \cite{mourikis_icra07, leutenegger, li_icra13, hesch_tro14, liu2017high, hesch_ijrr14, tanskanen_iros15, usenko_icra16, concha, jones2007inertial}. These approaches require tightly integrated sensor fusion, which places requirements for the hardware. For example, the synchronization of individual video frames and IMU sensor timestamps must be relatively accurate, and often the used IMUs are of notably better quality than standard smartphone IMUs, which are not aimed for inertial navigation purposes. In fact, many of the previous approaches utilize special hardware setups. For instance, both \cite{leutenegger} and \cite{usenko_icra16} use a similar synchronized IMU and stereo camera hardware prototype. Further, also \cite{hesch_tro14,hesch_ijrr14,concha,jones2007inertial} use custom-made camera-IMU hardware. Finally, perhaps the most well-known example of a specialized hardware platform for visual-inertial odometry is the Google Tango tablet device which utilizes a fish-eye lens camera \cite{tango}.  
 Regarding Google Tango device, it should be noticed that the implementation is proprietary and not openly documented, and hence it is difficult to analyze whether similar performance could be realized with more conventional smartphone hardware.

Nevertheless, there are some approaches which utilize standard smartphone sensors for motion tracking and metric reconstruction \cite{li_icra13,tanskanen_iccv13}. In \cite{tanskanen_iccv13} the authors report that the recovered scale was estimated to have an error of up to 10-15\% which is a significantly larger error than what we report in this paper. On the other hand, \cite{li_icra13} focus on real-time visual-inertial odometry in the navigation context, and not on precise metric reconstruction of objects like \cite{tanskanen_iccv13}. Thus, \cite{li_icra13} does not present a quantitative evaluation of the obtained scale accuracy and, as their implementation is not publicly available, detailed comparisons are not possible. Interestingly, although \cite{li_icra13} aims at real-time tracking, the authors report that in their experiments the data was recorded and stored onboard a phone and then processed \emph{offline} on the same device (but the average processing time per frame was small enough for real-time operation). Thus, it seems that there are not many publicly documented visual-inertial odometry systems that would be both accurate and capable of truly real-time operation on a smartphone.

Besides placing specific requirements for the hardware, tightly integrated fusion of visual and inertial measurements is a challenging task and leads to relatively complex designs as one needs to solve two difficult problems, visual odometry and inertial navigation, simultaneously. We believe that this complexity partially explains why many of the aforementioned state-of-the-art visual-inertial odometry methods are not available as open-source implementation.

In contrast to the approaches that utilize inertial sensors, purely visual reconstruction approaches are more mature and many solutions are available, both as open-source \cite{vsfm, openMVG, bundler, openSfM, lsdslam,dso} and commercial software \cite{pix4d,acute3d,capturingreality}. In particular, given the good visual quality and high accuracy and completeness of the reconstructed models that some of the commercial software packages provide, it appears that the only feature missing from these solutions is the absolute metric scale, which can not be obtained using images alone but which would be needed for certain measurement and modeling applications (e.g.\ body size measurements for clothing or reproduction of objects via 3D printing). 

Therefore, instead of aiming at tight integration of visual and inertial measurements for real-time odometry, we argue in favour of a decoupled approach, where one may use any visual structure-from-motion tool for capturing a 3D reconstruction from a smartphone video, and thereafter determine the metric scale of the reconstruction by applying the proposed approach for the inertial measurements, which were recorded simultaneously with the video capture. In fact, our experiments show that the proposed approach provides accurate results even when the precise temporal and spatial alignment between the video and IMU signals is not known a priori, or the camera poses provided by structure-from-motion are noisy and inaccurate (e.g.\ due to motion blur or rolling shutter effects). Further, besides having a wider application potential due to milder hardware requirements than most of the tightly integrated visual-inertial solutions, we believe that our batch-based approach has also potential for better accuracy thanks to the fact that it is able to utilize all the noisy IMU measurements for estimating the scale factor correction. Thus, if online or real-time scale estimation is not necessary, one may get a reasonably accurate scale even from low quality IMUs of commodity devices.

The closest previous works to ours are the papers by Ham et al.\ \cite{ham_eccv14,ham} which address the same problem in a similar context. That is, they also apply off-the-shelf visual tracking software to recover the camera poses up to scale and thereafter fix the metric scale based on inertial measurements. However, in contrast to Ham's approach we do not assume that the relative orientation of the camera and IMU must be known a priori. Further, we propose scale estimation by matching accelarations from visual and inertial sensors in frequency domain instead of time domain. We compare our frequency-based approach both to Ham's original implementation and to our own implementation of Ham's time domain approach. The results show that our approach has better accuracy and faster convergence of the scale estimate.

\section{BACKGROUND}
\subsection{Measurements}

Structure from motion software such as VisualSFM \cite{vsfm} outputs the camera poses for each image in the sequence. The pose defines the camera's position $\pv^{V}_W(t)$ and orientation $\Rm^{V}_W(t)$ in the world coordinate frame. Subscripts $W$ and $C$ denote the world and camera frames, respectively. The IMU measures the acceleration $\av^{I}_S(t)$ and angular rate $\omegav^{I}_S(t)$ of the device in the sensor coordinate frame, which is denoted with subscript $S$. Note that we use different superscripts for the visual and inertial measurements to avoid confusion.

\subsection{Transformations}

The orientation of the device can be represented by a rotation matrix $\Rm^{V}_W(t)$. To transform a vector in the world coordinates, e.g. the gravity vector $\gv_W$ to the camera coordinate frame, we apply the rotation
\begin{equation}
\label{WorldToCamera}
\gv_C(t) = \Rm^{V}_W(t) \; \gv_W.
\end{equation}

Assuming that the camera and the IMU are part of the same rigid structure and close together, there exists an orthogonal transformation $\Rm_S$ that rotates the inertial measurements to the camera coordinate frame
\begin{eqnarray}
\label{IMUToCamera}
&& \av^{I}_C(t) = \Rm_S\; \av^{I}_S(t) \\
&& \omegav^{I}_C(t) = \Rm_S\; \omegav^{I}_S(t). \nonumber
\end{eqnarray}

In Section \ref{sec:SpatialAlignment}, we propose a method for finding this rotation when the Euclidean transformation between the camera and IMU is unknown. It can be noted that the centripetal accelerations caused by rotations are assumed to be negligible small due to the close distance between the camera and the IMU.

\subsection{Overview of the algorithm}

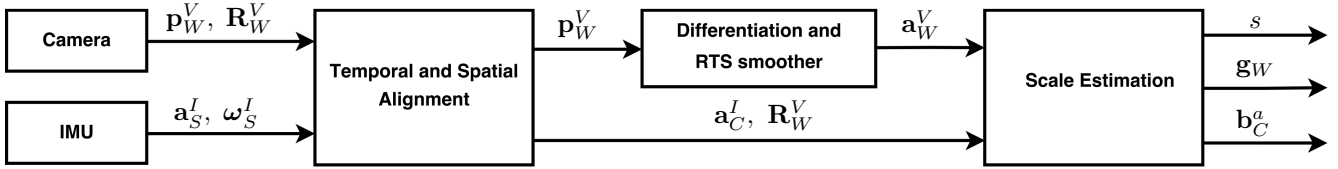
\begin{figure*}[t]
\begin{overpic}[width=1.0\textwidth]{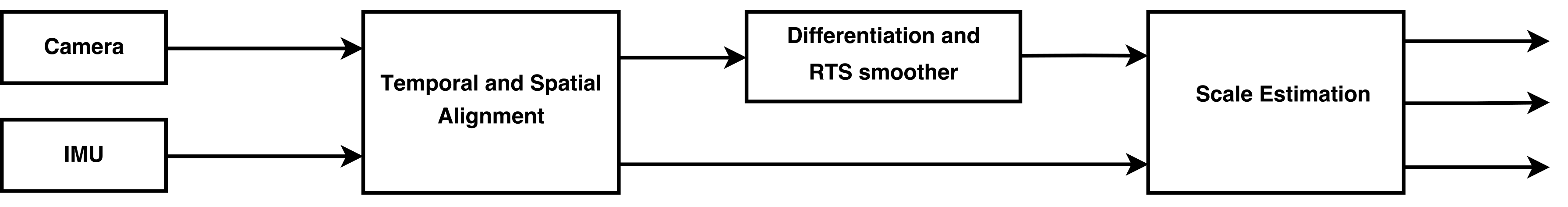}
 \put (12,11.3) {$\pv^{V}_W,$}
 \put (16.5,11.3) {$\Rm^{V}_W$}
 \put (12.6,4.2) {$\av^{I}_S,$}
 \put (16.3,4.2) {$\omegav^{I}_S$}
 \put (41.3,10.8) {$\pv^{V}_W$}
 \put (67,10.8) {$\av^{V}_W$}
 \put (53,4.0) {$\av^{I}_C,$}
 \put (57,4.0) {$\Rm^{V}_W$}
 \put (93,11.1) {$s$}
 \put (92,7.8) {$\gv_W$}
 \put (92,3.6) {$\bv^a_C$}
\end{overpic}
\caption{Processing steps of the proposed algorithm.}
\label{fig:2}
\end{figure*}

Processing steps of the proposed algorithm are visualized in Figure \ref{fig:2}. Before the scale estimation, we align the camera and IMU measurements both temporarily and spatially. This is achieved by comparing gyroscope readings and visual angular velocities, which are computed from the camera orientations. The scale estimation itself is performed by matching visual and inertial accelerations. More specifically, we differentiate the camera positions and match the accelerations in the frequency domain. Structure from motion algorithms typically assume that the input is a collection of unordered images. Thus, the continuity of motion is ignored when reconstructing from a video. This may lead to noisy position estimates reducing the accuracy of the scale estimate. Since we can expect that the device follows physical laws of motion, we employ the Rauch-Tung-Striebel (RTS) smoother to refine the position estimates.

\section{VISUAL-IMU ALIGNMENT}
The scale estimation relies on the assumption that the visual and inertial measurements are temporally aligned. Furthermore, we want to ensure that the camera coordinate frame is spatially aligned with the coordinate frame of the inertial measurement unit.

\subsection{Temporal alignment}
\label{sec:TemporalAlignment}

The timestamps of the IMU and visual-data are a natural starting point for temporal alignment. The camera timestamps are not always available, and even when they are, the camera and IMU may be using different clocks. This prevents the direct comparison of the measurements since there is an unknown temporal offset between the timestamps. Both the literature and our experiments confirm that the mapping between the IMU and visual-data is not just a constant shift operation. The time offset can slightly change over time, due to inaccuracies in the sensors' clocks, or clock jitters from CPU overloading \cite{mair}.

\begin{figure}[t]
\includegraphics[width=0.5\textwidth]{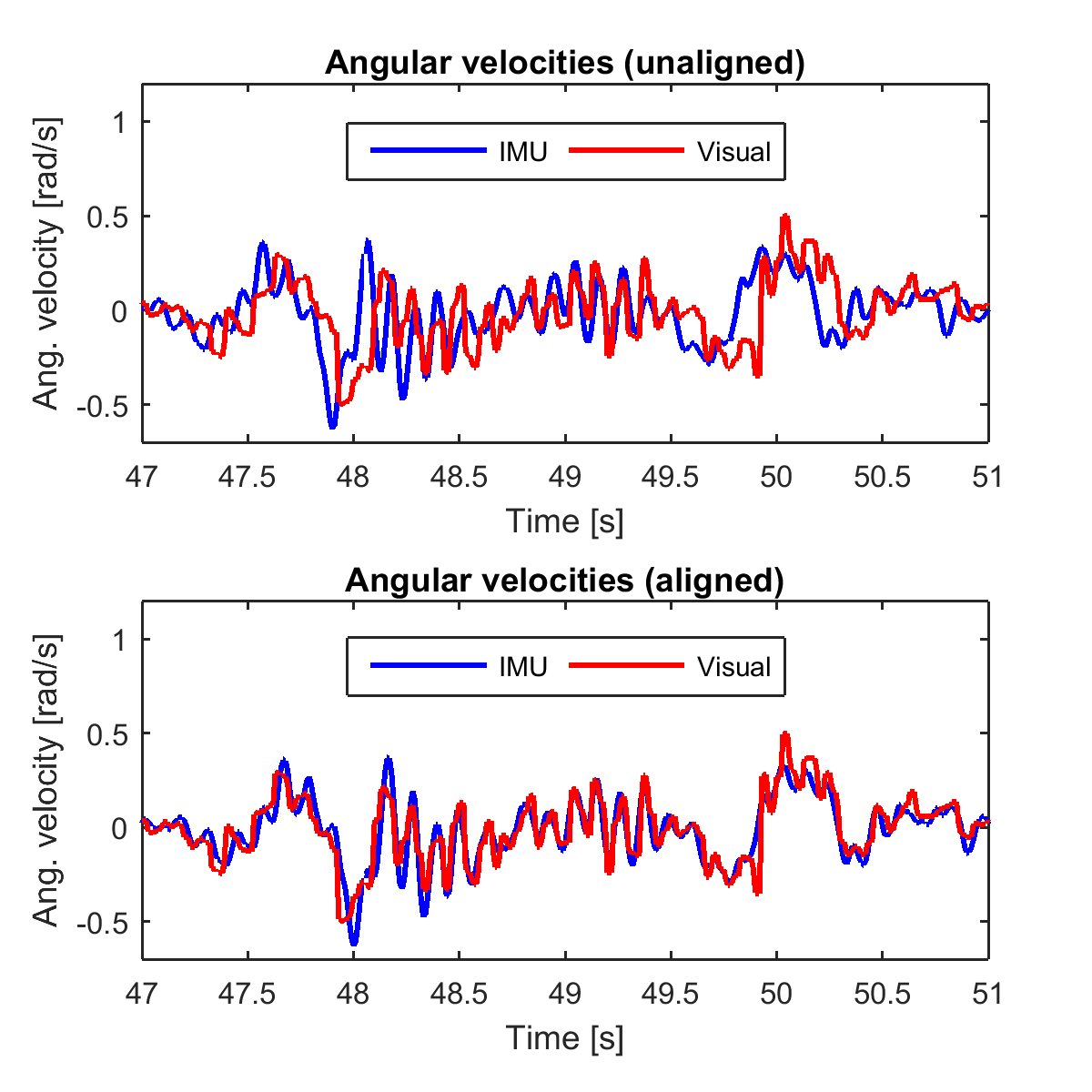}
\caption{Comparison of visual and inertial angular velocities before and after the temporal alignment.}
\label{fig:3}
\end{figure}

Our solution to temporal alignment is based on the idea of comparing gyroscope readings with the visual angular velocities, which are computed from the camera orientations. The offset value, which minimizes the least-squares error is chosen as the best estimate. We perform the temporal alignment concurrently with the spatial alignment since the angular velocities cannot be compared without knowing the spatial alignment between the coordinate frames. This topic is discussed in Section \ref{sec:SpatialAlignment}. Figure \ref{fig:3} shows the angular velocities before and after the temporal alignment. Note that the angular velocities have been spatially aligned for visualization.

Previous methods, such as \cite{ham} assume a constant offset but as mentioned, the offset can slightly vary over time. In fact, we can see from Figure \ref{fig:3} that signals are not perfectly aligned after compensating for a constant delay. Even though the difference is barely noticeable, we can take this account by performing the scale estimation in frequency domain as will be shown in Section \ref{sec:frequencyDomain}.

\subsection{Spatial alignment}
\label{sec:SpatialAlignment}

The rotation matrix $\Rm_S$, which aligns the sensor coordinate frame with the camera coordinate frame may often be deduced from the documentations of the development platform and the reconstruction software. There is, however, no guarantee that such transformation is precisely correct. Our approach does not require that the transformation is known in advance. Instead, we estimate the transformation by utilizing gyroscope measurements. The aim is to find the optimal rotation between two sets of angular velocities. First we compute the visual angular velocities $\omegav^{V}_C(t)$ from the camera orientations $\Rm^{V}_W(t)$. The relation between the angular velocity and the derivative of the rotation matrix is given by the equation
\begin{equation}
[\omegav^{V}_C]_{\times} = \frac{d \, \Rm^{V}_W}{dt} \Rm^{V^{\top}}_W.
\end{equation}
The components of the angular velocity can be extracted from the 3 $\times$ 3 skew-symmetric matrix $[\omegav^{V}_C]_{\times} $.

We can find the transformation between the angular velocities $\omegav^{V}_C(t)$ and $\omegav^{I}_S(t)$ by minimizing
\begin{equation}
\label{eq_minRT}
\underset{\Rm_S, \bv^{\omega}_C}{\argmin} \sum_{t} \| \omegav^{V}_C(t) - (\Rm_S\; \omegav^{I}_S(t) + \bv^{\omega}_C) \|^2,
\end{equation}
where $\bv^{\omega}_C$ is the translation vector, which equals to gyroscope bias in the camera coordinate frame. It has been shown \cite{arun,kanatani} that this problem can be solved optimally in closed-form. Indeed, we may first translate the angular velocities so that their centroids are at the origin of the coordinate frame and then solve the optimal rotation $\Rm_S$, and thereafter $\bv^{\omega}_C$, as in \cite{arun}.

Thus far, we have assumed that the visual and inertial angular velocities are temporally aligned. Since this may not be true, we include the offset term $t_d$ to the objective function
\begin{equation}
\label{eq_minRoffset}
\underset{\Rm_S, \bv^{\omega}_C, t_d}{\argmin} \sum_{t} \| \omegav^{V}_C(t) - (\Rm_S\; \omegav^{I}_S(t+t_d) + \bv^{\omega}_C) \|^2,
\end{equation}
which is minimized using alternating optimization. That is, we find the optimal $t_d$ iteratively using golden section search, where we solve $\Rm_S$ and $\bv^{\omega}_C$ in closed-form at each iteration. 

Before the scale estimation, we align the inertial accelerations to the camera frame both temporally and spatially using the estimated values for $\Rm_S$ and $t_d$ (cf.\ Figure \ref{fig:2}).

\section{SCALE ESTIMATION}
The aim of the scale estimation is to find a scale factor that fixes the metric scale of the reconstruction. Our work is similar to \cite{ham} in the sense that we compare accelerations instead of positions. What makes the problem more complicated is the fact that the accelerometer also measures the earth's gravity, which is not observed by the camera. Furthermore, the measurements are corrupted by the noise and the accelerometer readings may be biased. In practice, we not only estimate the scale but the gravity and accelerometer bias as well.


\subsection{Visual accelerations and RTS smoothing}

\begin{figure}[t]
\includegraphics[width=0.5\textwidth]{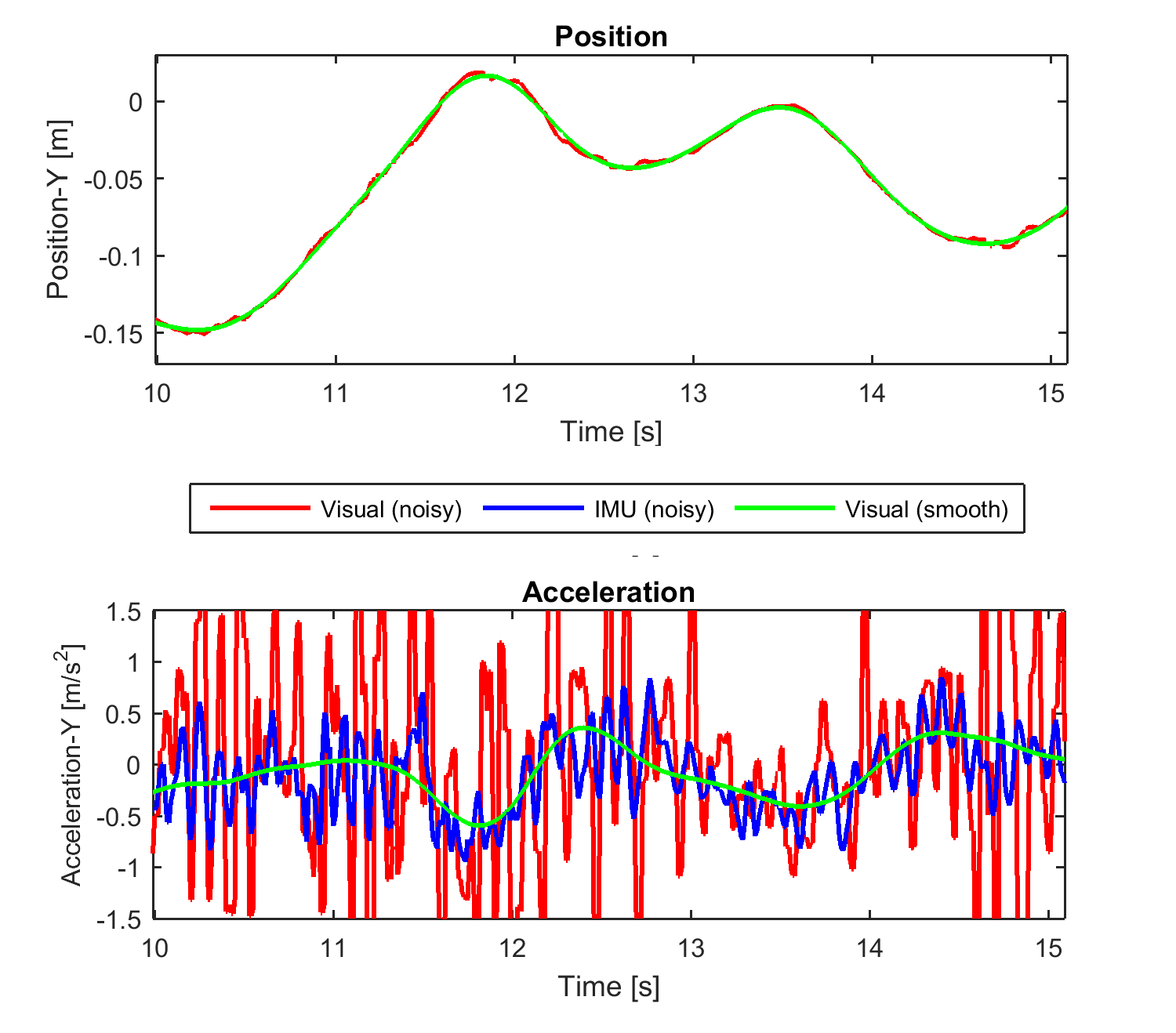}
\caption{Visual positions and accelerations before and after RTS smoothing.}
\label{fig:4}
\end{figure}

Assuming that we know the time interval between each frame in the video, we can compute the visual accelerations by taking the second derivative of the position. Figure \ref{fig:4} shows the original camera positions and the corresponding accelerations. As can be seen, the differentiation amplifies the noise in the original signal. Using these noisy accelerations in the scale estimation would cause the scale to be severely underestimated.


The method in \cite{ham} addresses this problem by applying a low-pass filter to the visual and inertial accelerations. Based on our experiments, the choice of the cutoff frequency has significant impact to the accuracy of the scale estimation. The difficulty is that some sequences may require more smoothing than others. The amount of noise may also vary within different parts of the sequence.



Kalman filter \cite{Kalman:1960} can also be used to filter position data. Its advantage is that it takes into account that the device is expected to follow physical laws of motion. The algorithm weights the measurements and predicted states based on their uncertainties. However, it does not, as such, solve the problem of varying amounts of smoothing required by the sequences.

In order to cope with the requirement for varying amount of smoothing, we adapt the process noise parameters of the Kalman filter state space model by using the marginal maximum likelihood method (e.g., \cite{Sarkka:2013}). We compute likelihood curve on a one-dimensional grid and automatically select the best process noise value based on the marginal likelihood obtained by the prediction error decomposition.

To further improve the estimates, we employ the Rauch-Tung-Striebel (RTS) smoother \cite{rauch}. Unlike the Kalman filter, the RTS smoother also utilizes the future samples to determine the optimal smoothing. It is a two-pass algorithm for fixed-interval smoothing, where the forward pass corresponds to the Kalman filter. The state estimates and covariances are stored for the backward pass. Figure \ref{fig:4} shows the smoothed positions and accelerations, which are computed in the backward pass. 

\subsection{Scale and bias estimation}

When comparing visual and inertial accelerations we have to take into account that the accelerometer not only measures the acceleration caused by the motion but also the acceleration due to earth's gravity. Most smart device APIs provide estimates of the linear acceleration in addition to the raw accelerometer readings. In case of linear acceleration, the scale estimation is simpler because the accelerometer readings do not contain the gravity component. Later on, we show that we do not have to rely on the built-in black box gravity estimation but let us first consider a case of linear acceleration. Given the visual accelerations $\av^{V}_C(t)$ and the linear accelerometer readings $\av^{I}_C(t)$, the scale factor $s$ can be estimated by minimizing
\begin{equation}
\label{eq:mins}
\underset{s}{\argmin} \sum_{t} \| s \, \av^{V}_C(t) - \av^{I}_C(t) \|^2.
\end{equation}

Note that visual accelerations have been rotated to the camera coordinate frame using \eqref{WorldToCamera}. The inertial accelerations have also been aligned with the camera coordinate frame using \eqref{IMUToCamera}.

The cost function \eqref{eq:mins} disregards the fact that accelerometer readings are typically biased. Let us denote the accelerometer bias in the camera coordinate frame with $\bv^a_C$. It is known that the bias may depend on the temperature of the sensor \cite{aggarwal}. According to our experiments, we may assume that the bias stays constant during the recording. We also experimented with the bias term that varies linearly with the time but did not observe any noticeable improvement. The objective function with the constant bias can be written as
\begin{equation}
\label{eq:minsb}
\underset{s,\bv^a_C}{\argmin} \sum_{t} \| s \, \av^{V}_C(t) - \av^{I}_C(t) + \bv^a_C \|^2.
\end{equation}

\subsection{Gravity estimation}

As mentioned, the gravity vector is included in the raw accelerometer readings. In order to compare these accelerations, we need to subtract the gravity component from the IMU measurements. Since the gravity vector $\gv_W$ is constant in the world coordinate frame, we can transform this vector to the camera coordinate frame using \eqref{WorldToCamera}. Therefore, we only need to estimate the three parameters of the gravity vector. The objective function with the gravity term can be written as
\begin{gather}
\underset{s,\bv^a_C,\gv_W}{\argmin} \sum_{t} \| s \, \av^{V}_C(t) - \av^{I}_C(t) + \bv^a_C + \Rm^{V}_W(t) \gv_W \|^2, \nonumber \\ 
\text{subject to} \; \| \gv_W \|^2 = 9.8.
\label{eq:minsbg}
\end{gather}

Note that the system is linear but there is a nonlinear gravity constraint that needs to be satisfied. In practice, the accuracy of the scale seems to be almost as good even if the
gravity constraint is ignored.

\subsection{Frequency domain representation}
\label{sec:frequencyDomain}

In this section, we show that the scale estimation can also be performed in the frequency domain. This representation has clear advantages over the time domain as will be demonstrated in the experiments. It was discussed earlier, that the visual accelerations need to be smoothed for accurate results. The frequency domain representation allows us to simply disregard the high frequency components, i.e.\ the noise in the visual and inertial accelerations. This way we can avoid the difficulty of choosing the right cutoff frequency for the low-pass filter. Another advantage of the frequency domain representation is its robustness against the phase difference between the inertial and visual data.  As discussed in Section \ref{sec:TemporalAlignment}, the temporal offset of the inertial and visual measurements may slightly vary over time. This may cause problems when the scale estimation is performed in the time domain.

Let us denote the Fourier transform with operator $\mathcal{F} \{ \cdot \}$. From \eqref{eq:minsbg}, we get the visual and inertial accelerations
\begin{gather}
\label{eq:fourier}
\Am^{V}(f) = \mathcal{F} \{ s \, \av^{V}_C(t) \}, \\
\Am^{I}(f) = \mathcal{F} \{ \av^{I}_C(t) - \bv^a_C - \Rm^{V}_W(t) \gv_W \},
\end{gather}
where $\Am^{V}(f)$ and $\Am^{I}(f)$ are matrices that contain the Fourier transforms of the visual and inertial accelerations, respectively. Note that the transform is taken separately for each of the three axis. We wish to minimize the difference between the amplitude spectrums $| \Am^{V}(f) |$ and  $| \Am^{I}(f) |$
\begin{gather}
\underset{s,\bv^a,\gv_W}{\argmin} \sum^{f_{max}}_{f} \| \; | \Am^{V}(f) | - | \Am^{I}(f) | \; \|^2, \nonumber \\ 
\text{subject to} \; \| \gv_W \|^2 = 9.8.
\label{eq:minfreq}
\end{gather}

\begin{figure}[t]
\includegraphics[width=0.5\textwidth]{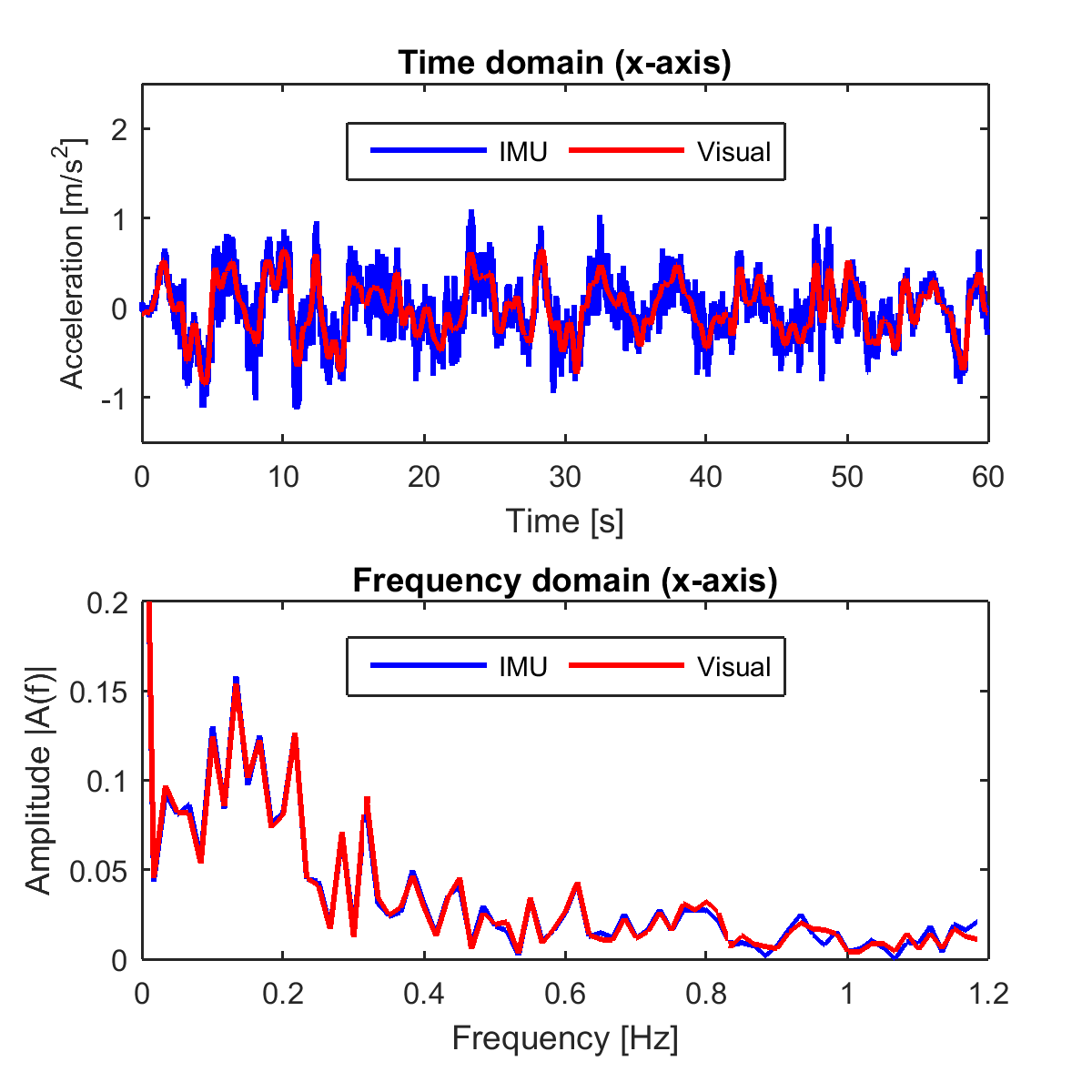}
\caption{Time and frequency domain representations of the visual and inertial accelerations. In both cases, the visual accelerations have been scaled with the estimated value.}
\label{fig:5}
\end{figure}

The upper limit for the frequencies is denoted by $f_{max}$. In our experiments, this value was set to 1.2 Hz. We did not find it necessary to tune the parameter. To ensure a good initialization for the minimization, we use the closed-form solution of \eqref{eq:minsbg} as an initial guess (i.e.\ we dropped the nonlinear constraint in \eqref{eq:minsbg} but not in \eqref{eq:minfreq}). The above problem is then solved with the \emph{fmincon} function provided by the optimization toolbox of Matlab. Figure \ref{fig:5} shows the time and frequency domain representations of the inertial and visual accelerations.

\begin{figure*}[t]
 \includegraphics[width=1\textwidth]{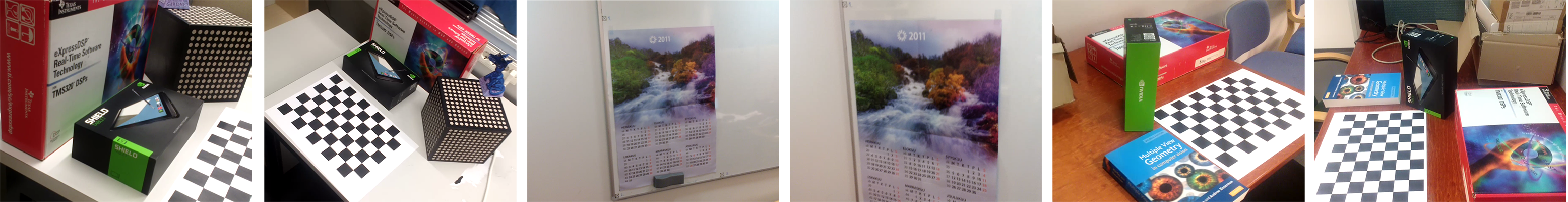}
\caption{Individual frames from the static test sequences.}
\label{fig:6}
\vspace{-3mm}
\end{figure*}

\section{EVALUATION}
We evaluate our method on several datasets captured with the NVIDIA Shield tablet \cite{shield} and Project Tango Development Kit \cite{tango}. The main difference between these devices is that the Tango has a build-in motion tracking system, which provides pose estimates in real-time. In case of the NVIDIA Shield, the camera poses were obtained using the VisualSFM software \cite{vsfm}. We compare our solution against the state-of-the-art method proposed by Ham et al. \cite{ham} and the Tango's build-in motion tracking.

The method \cite{ham} has a parameter that determines the amount of smoothing applied to the accelerations. This parameter was here tuned so that the average error of the scale with respect to the ground truth was lowest possible. It can be noted that this may give over-optimistic results since the same datasets were used in training and evaluation. 

It should be also noted that our method does not have parameters that would need to be manually adjusted per sequence. That is, all parameters are either set automatically or kept at a fixed value for all test sequences.

Further, in contrast to our method, the method \cite{ham} needs the IMU-to-camera transformation as an input. The transformation was set as defined in Android documentation.

Finally, although our implementation is not optimized for speed, it is clear that the scale estimation is computationally light compared to the visual reconstruction. For example, processing the $60s$ long test sequences took less than $1s$ on a laptop with our Matlab implementation.

\subsection{Static scene experiments}

In these experiments, we captured 15 datasets with the NVIDIA Shield tablet. The IMU was logging at 100 Hz, which is the fastest possible rate for this device.  The camera was recording at 30 fps and with resolution of 1080x1920 pixels. The camera pose trajectory was upsampled to match the sample rate of the IMU. The timestamps were available for both measurements, although the camera and IMU were using different clocks.  The camera poses were obtained using the VisualSFM software. The ground truth scale was determined using the software's build-in feature for ground control point registration. For this purpose, we embedded a checkerboard pattern (or other known 3D points) to the scene.

Figure \ref{fig:6} shows individual frames from the datasets. 
 The objects in the scenes are viewed from different distances and angles. In general, the camera motion is not particularly fast but many of the video frames clearly suffer from motion blur. Each sequence is 60 seconds long and the total distance the camera travels varies between 14 and 20 meters. In the experiments, we wanted to make sure that there is no visual drift. Therefore, at least some parts of the scene remain visible the entire length of the recording. In case of visual drift, the scale of the reconstruction might change over time.

\begin{figure}[t]
 \includegraphics[width=0.5\textwidth]{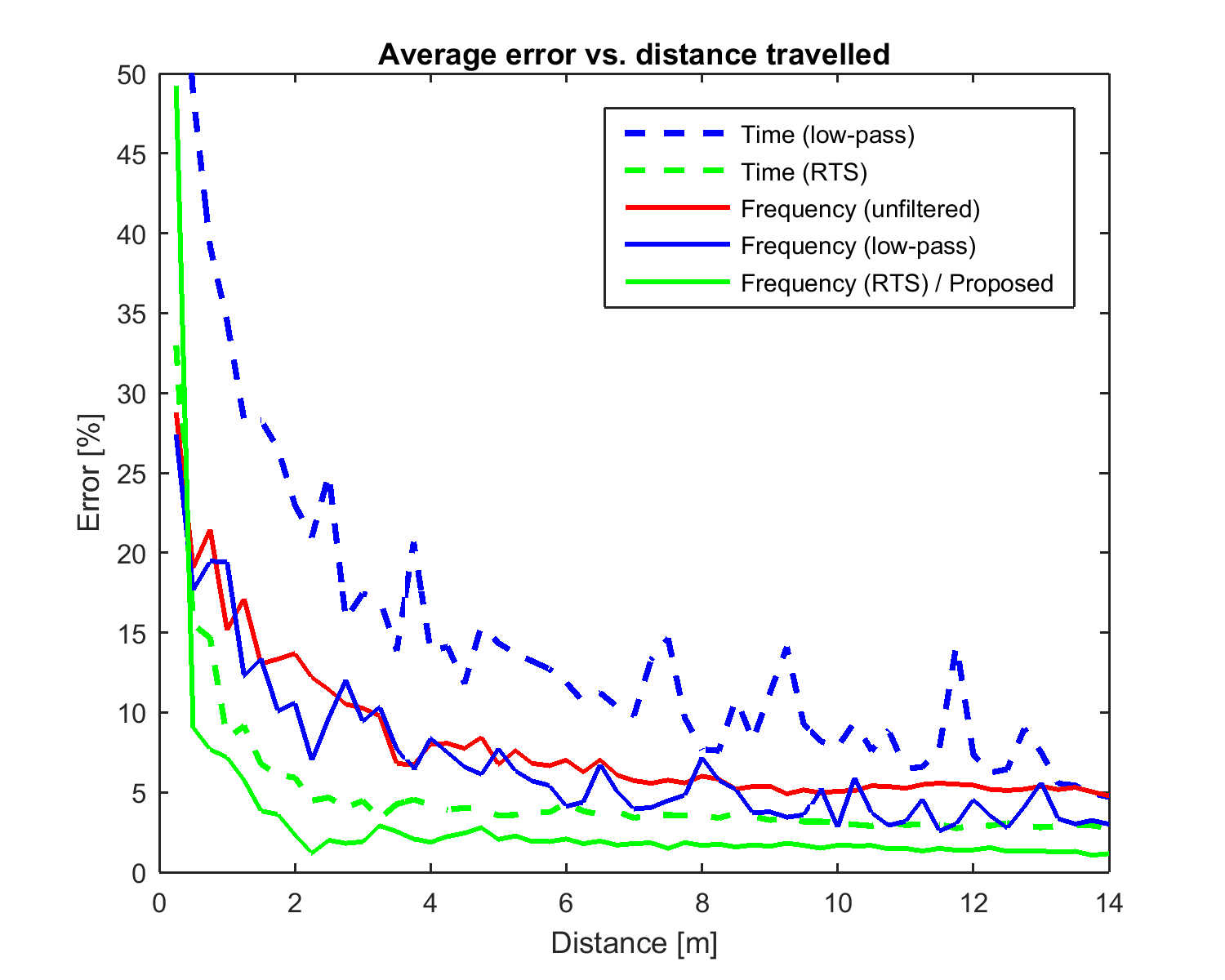}
\caption{Comparison of time and frequency domain representations while using different methods for smoothing.}
\label{fig:7}
\end{figure}

In practice, we only need to run the scale estimation once using all the samples available. However, we are also interested how quickly the scale converges towards the true value. The green line in Figure \ref{fig:7} shows the average error of the scale after the camera has traveled a certain distance. We can clearly see that the accuracy improves the farther the camera travels. The average error of the scale is already below 3 $\%$ after the camera has traveled 2 meters.

A comparison of the RTS smoother and a low-pass filter is show in Figure \ref{fig:7}. Here we also compare the frequency domain method against the time domain representation. We can see that the RTS smoother outperforms the low-pass filter, regardless of the method used and distance traveled. The best results are obtained when the RTS smoother is used together with the frequency domain method. As a reference, the graph also shows the results when smoothing is not applied. In such case, the error of time domain method is near 90 percent while the frequency domain method is still able to estimate the scale around 5 $\%$ accuracy.

\begin{figure}[t]
\includegraphics[width=0.5\textwidth]{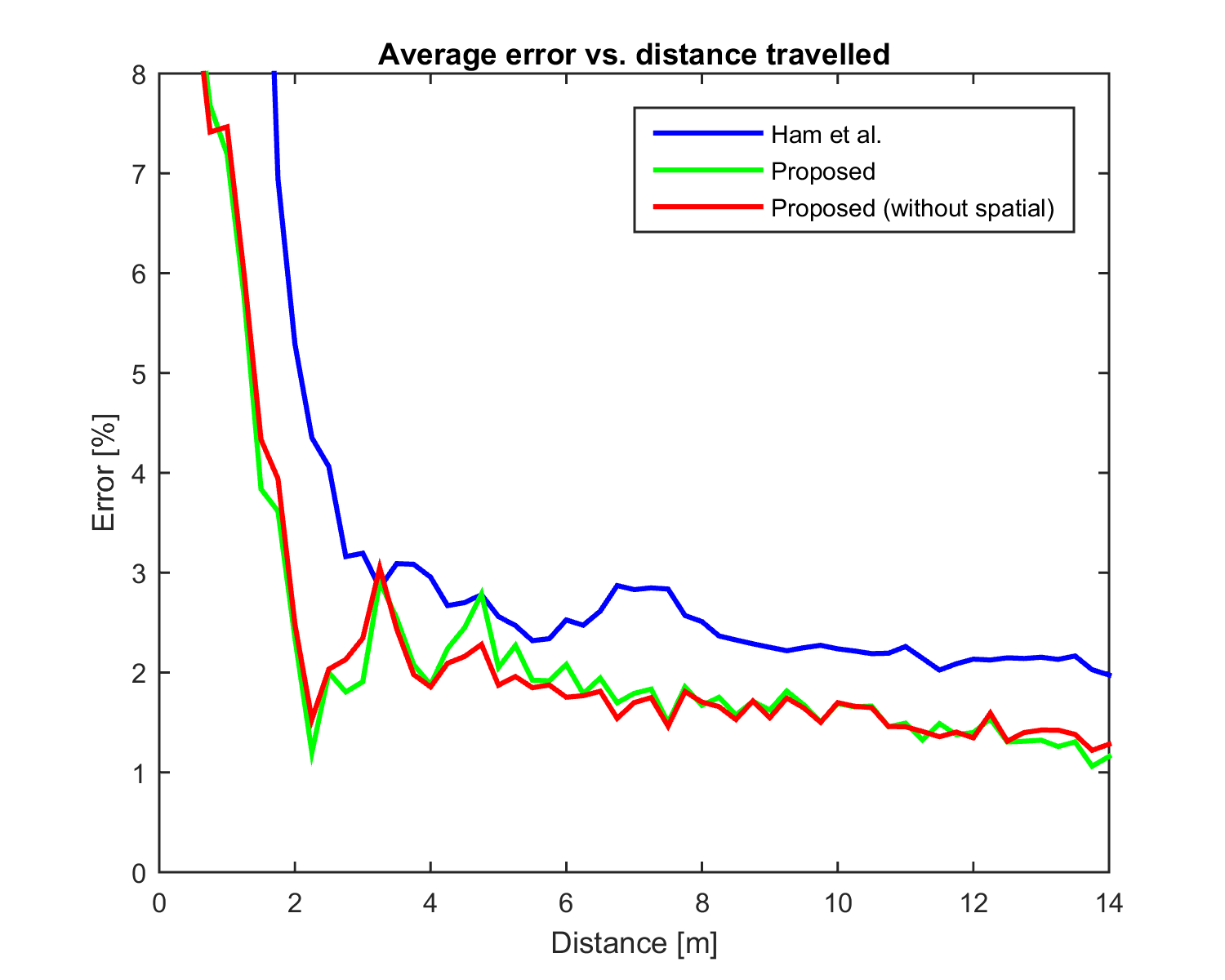}
\caption{Comparison of our method against Ham et al.\ \cite{ham}}
\label{fig:8}
\end{figure}

We compare our method against the algorithm proposed by Ham et al.\ \cite{ham}. Table \ref{tab:1} summarizes the results for individual datasets. The average error is visualized in Figure \ref{fig:8}. With our method, the average error is around 1 $\%$ after the camera has traveled 14 meters. The corresponding error for the Ham et al. is around 2 $\%$. We can also see that our algorithm converges more quickly towards the true scale.

In the previous experiments, the spatial alignment of the camera and IMU was estimated from the input data. For comparison, Figure \ref{fig:8} shows the average errors in case the spatial alignment is defined in advance. Here we have used the same IMU-to-camera calibration matrix as with the algorithm \cite{ham}. The results show that the accuracy of the scale is equally good when the transformation is estimated from the input data. This is true even when the sequence is short.



\begin{table}
\newcolumntype{.}{D{.}{.}{-1}}
\centering
\setlength{\tabcolsep}{3pt}
\caption{Error of the scale (\%) in relation to distance traveled.}
\begin{tabular}{l....@{\hskip 16pt}....}
\toprule
\multicolumn{1}{c}{} &
\multicolumn{4}{c}{Ham \cite{ham}} &
\multicolumn{4}{c}{Proposed} \\
\cmidrule(r){2-5}
\cmidrule(r){6-9}
\multicolumn{1}{l}{Dataset} & 
\multicolumn{1}{c}{1m} &
\multicolumn{1}{c}{2m} &
\multicolumn{1}{c}{6m} &
\multicolumn{1}{c}{14m} &
\multicolumn{1}{c}{1m} &
\multicolumn{1}{c}{2m} &
\multicolumn{1}{c}{6m} &
\multicolumn{1}{c}{14m} \\
\cmidrule(r){2-5}
\cmidrule(r){6-9}
\#1  & 40.8 &   4.9 &   2.9 &   0.2 &      6.0 &   7.6 &   2.1 &   0.5 \\
\#2  &  8.0 &   6.2 &   3.7 &   2.1 &      1.2 &   3.1 &   0.2 &   0.7 \\
\#3  & 10.9 &   2.9 &   0.2 &   0.1 &      2.0 &   1.7 &   0.2 &   0.5 \\
\#4  & 18.0 &   4.5 &   3.9 &   2.0 &      0.8 &   4.3 &   0.9 &   0.6 \\
\#5  & 15.6 &   2.5 &   3.0 &   1.1 &      1.3 &   0.3 &   0.6 &   1.0 \\
\#6  & 18.6 &  11.2 &   1.8 &   0.2 &      4.6 &   1.4 &   5.5 &   0.2 \\
\#7  &  3.2 &   2.9 &   2.4 &   3.0 &     10.2 &   1.4 &   0.0 &   1.6 \\
\#8  & 74.8 &   4.2 &   0.7 &   0.8 &      6.7 &   1.8 &   0.1 &   0.8 \\
\#9  & 58.3 &   1.2 &   1.4 &   1.2 &     11.6 &   3.7 &   1.4 &   2.7 \\
\#10 & 94.8 &   2.3 &   1.5 &   0.0 &     26.3 &   2.2 &   7.9 &   0.2 \\
\#11 &  2.0 &   7.6 &   3.2 &   2.3 &      0.9 &   0.5 &   0.1 &   0.4 \\
\#12 & 13.0 &  12.8 &   3.0 &   2.9 &      1.1 &   2.0 &   2.6 &   2.0 \\
\#13 & 18.6 &   8.4 &   4.9 &   6.0 &      4.1 &   2.8 &   2.3 &   3.5 \\
\#14 &  7.1 &   3.2 &   4.2 &   4.6 &      5.7 &   1.3 &   3.7 &   2.0 \\
\#15 &  2.1 &   3.8 &   0.4 &   2.4 &      3.3 &   0.5 &   2.9 &   0.0 \\
\bottomrule
\end{tabular}
\label{tab:1}
\vspace{-3mm}
\end{table}

\subsection{Height measuring experiments}
Here we intend to find out how accurately can we estimate the height of a person using our method. Unlike in the previous experiments, the scene is not completely static. The person was, however, advised to remain stationary during the recording.
Datasets were captured with the NVIDIA Shield table and they consist of one indoor sequence (50 seconds) and one outdoor sequence (40 seconds).

Figure \ref{fig:1} shows the scaled point clouds. Most of the points in the background were removed for clarity, especially the points above the head and below the ground plane. The point clouds were also rotated so that the gravity vector is aligned with the y-axis.

For these experiments, we did not place any artificial objects to the scene (e.g. the checkerboard). Instead, the true height of the person was measured to be approximately 174 cm. The estimated heights were 176 cm (indoor sequence) and 181 cm (outdoor sequence). These values correspond to errors 1.1 \% and 4\%, respectively. It should be noted that factors such as the person's posture, clothing and accuracy of the physical height measurement will affect the results.

\subsection{Project Tango experiments}

The sequences in these experiments were recorded with the Project Tango Development Kit \cite{tango}. In contrast to previous experiments, there is no need to use any structure from motion software since the Tango provides pose estimates in real time at the rate of 100 Hz. Even though these estimates are in metric units, we intend to find out if our method can further improve the accuracy of the scale.

In order to evaluate the accuracy of the Tango's scale, we set up a test room with known ground point locations. While recording, the device was briefly stopped at each of the four locations and the corresponding points were extracted from the Tango's trajectory. The points were then rotated and translated to fit the ground points. The optimal transformation was estimated using the algorithm \cite{arun}. After the fitting, the root-mean-square error (RMSE) was computed to evaluate the goodness of the fit. To find out if the results can be improved by our method, we scaled the estimated points with the estimated scale factor and fit the points again.

Table \ref{tab:2} shows the RMS-errors between the ground points and the estimated points, with and without the scale adjustment. Even though the Tango's scale is already quite accurate, we can see that our method improves the results in almost every case. A common observation for all the Tango experiments is that the estimated scale factor is always slightly above one. This indicates that in these experiments, the Tango tends to underestimate the scale. In fact, the reason why the error sometimes increases is not caused by the inaccurate scale estimate but rather the fact that Tango's motion tracking sometimes clearly fails. In such a case, the error cannot be reduced by simply adjusting the scale, because the pose tracking suffers also from other errors.

\begin{table}
\caption{Root-mean-square errors (in meters) between the camera positions and ground points with and without the scale adjustment.}
\label{tab:2}       
\setlength{\tabcolsep}{13pt}
\begin{tabular}{@{\hspace{0\tabcolsep}}lcc}
\hline\noalign{\smallskip}
{\normalsize Dataset} & {\normalsize RMSE (build-in)} & {\normalsize RMSE (proposed)}  \\
\noalign{\smallskip}\hline\noalign{\smallskip}
{\normalsize \#1} & {\normalsize 0.0087} & {\normalsize 0.0074} \\
{\normalsize \#2} & {\normalsize 0.0201} & {\normalsize 0.0176} \\
{\normalsize \#3} & {\normalsize 0.0057} & {\normalsize 0.0055} \\
{\normalsize \#4} & {\normalsize 0.0364} & {\normalsize 0.0392} \\
{\normalsize \#5} & {\normalsize 0.0272} & {\normalsize 0.0249} \\
{\normalsize \#6} & {\normalsize 0.0145} & {\normalsize 0.0146} \\
{\normalsize \#7} & {\normalsize 0.0176} & {\normalsize 0.0141} \\
{\normalsize \#8} & {\normalsize 0.0126} & {\normalsize 0.0074} \\
{\normalsize \#9} & {\normalsize 0.0223} & {\normalsize 0.0178} \\
{\normalsize \#10} & {\normalsize 0.0075} & {\normalsize 0.0057} \\
\noalign{\smallskip}\hline
\end{tabular}
\end{table}

\section{CONCLUSION}
We have proposed a method which recovers the metric scale of a visual structure-from-motion reconstruction by utilizing inertial measurements. The algorithm can be easily bundled with existing reconstruction software since the spatial and temporal alignment of the camera and IMU does not have to be known in advance. The evaluation shows that our method outperforms the current state-of-the-art in both accuracy and convergence rate of the scale estimate. The accuracy of the scale estimate is typically around 1 \% depending on the distance traveled. We have also demonstrated that our method can improve the scale estimate of Project Tango's build-in motion tracking.

\bibliographystyle{IEEEtran}
\bibliography{IEEEabrv,bibfile}

\begin{thebibliography}{10}
\providecommand{\url}[1]{#1}
\csname url@rmstyle\endcsname
\providecommand{\newblock}{\relax}
\providecommand{\bibinfo}[2]{#2}
\providecommand\BIBentrySTDinterwordspacing{\spaceskip=0pt\relax}
\providecommand\BIBentryALTinterwordstretchfactor{4}
\providecommand\BIBentryALTinterwordspacing{\spaceskip=\fontdimen2\font plus
\BIBentryALTinterwordstretchfactor\fontdimen3\font minus
  \fontdimen4\font\relax}
\providecommand\BIBforeignlanguage[2]{{%
\expandafter\ifx\csname l@#1\endcsname\relax
\typeout{** WARNING: IEEEtran.bst: No hyphenation pattern has been}%
\typeout{** loaded for the language `#1'. Using the pattern for}%
\typeout{** the default language instead.}%
\else
\language=\csname l@#1\endcsname
\fi
#2}}

\bibitem{rulerphone}
``B. {Kamens}, {RulerPhone - Photo Measuring. Apple App Store},''
  \url{https://itunes.apple.com/au/app/rulerphone-camera-measuring/id288774794?mt=8},
  accessed: 2017-02-28.

\bibitem{thirdlove}
``{ThirdLove - Lingerie and Virtual Sizing Technology},''
  \url{https://www.thirdlove.com/pages/ios-app}.

\bibitem{smartmeasure}
``{Smart Tools co., Smart Measure Pro. Google Play Store},''
  \url{https://play.google.com/store/apps/details?id=kr.aboy.measure},
  accessed: 2017-02-28.

\bibitem{mourikis_icra07}
A.~I. Mourikis and S.~I. Roumeliotis, ``A multi-state constraint {Kalman}
  filter for vision-aided inertial navigation,'' in \emph{IEEE International
  Conference on Robotics and Automation (ICRA)}, 2007.

\bibitem{leutenegger}
S.~Leutenegger, S.~Lynen, M.~Bosse, R.~Siegwart, and P.~Furgale,
  ``Keyframe-based visual--inertial odometry using nonlinear optimization,''
  \emph{International Journal of Robotics Research}, vol.~34, no.~3, pp.
  314--334, 2015.

\bibitem{li_icra13}
M.~Li, B.~H. Kim, and A.~I. Mourikis, ``Real-time motion tracking on a
  cellphone using inertial sensing and a rolling-shutter camera,'' in
  \emph{IEEE International Conference on Robotics and Automation (ICRA)}, 2013.

\bibitem{hesch_tro14}
J.~A. Hesch, D.~G. Kottas, S.~L. Bowman, and S.~I. Roumeliotis, ``Consistency
  analysis and improvement of vision-aided inertial navigation,'' \emph{{IEEE}
  Trans. Robotics}, vol.~30, no.~1, pp. 158--176, 2014.

\bibitem{liu2017high}
T.~Liu and S.~Shen, ``High altitude monocular visual-inertial state estimation:
  Initialization and sensor fusion,'' 2017.

\bibitem{hesch_ijrr14}
J.~A. Hesch, D.~G. Kottas, S.~L. Bowman, and S.~I. Roumeliotis,
  ``{Camera-IMU-based localization: Observability analysis and consistency
  improvement},'' \emph{International Journal of Robotics Research}, vol.~33,
  no.~1, pp. 182--201, 2014.

\bibitem{tanskanen_iros15}
P.~Tanskanen, T.~{N\"ageli}, M.~Pollefeys, and O.~Hilliges, ``Semi-direct
  {EKF-based} monocular visual-inertial odometry,'' in \emph{IEEE/RSJ
  International Conference on Intelligent Robots and Systems (IROS)}, 2015.

\bibitem{usenko_icra16}
V.~Usenko, J.~Engel, J.~{St\"uckler}, and D.~Cremers, ``Direct visual-inertial
  odometry with stereo cameras,'' in \emph{IEEE International Conference on
  Robotics and Automation (ICRA)}, 2016.

\bibitem{concha}
A.~Concha, G.~Loianno, V.~Kumar, and J.~Civera, ``Visual-inertial direct
  {SLAM},'' in \emph{2016 IEEE International Conference on Robotics and
  Automation (ICRA)}.\hskip 1em plus 0.5em minus 0.4em\relax IEEE, 2016, pp.
  1331--1338.

\bibitem{jones2007inertial}
E.~Jones, A.~Vedaldi, and S.~Soatto, ``Inertial structure from motion with
  autocalibration,'' in \emph{Workshop on Dynamical Vision}, vol.~25, 2007.

\bibitem{tango}
``{Project Tango Development Kit},''
  \url{https://developers.google.com/tango/}, accessed: 2017-02-28.

\bibitem{tanskanen_iccv13}
P.~Tanskanen, K.~Kolev, L.~Meier, F.~Camposeco, O.~Saurer, and M.~Pollefeys,
  ``Live metric {3D} reconstruction on mobile phones,'' in \emph{International
  Conference on Computer Vision (ICCV)}, 2013.

\bibitem{vsfm}
C.~Wu, ``{VisualSfM, a visual structure from motion system},''
  \url{http://ccwu.me/vsfm/}, accessed: 2017-02-28.

\bibitem{openMVG}
P.~Moulon, P.~Monasse, R.~Marlet, and Others, ``{OpenMVG. An Open Multiple View
  Geometry library.}'' \url{https://github.com/openMVG/openMVG}.

\bibitem{bundler}
S.~Snavely, ``Bundler: structure from motion for unordered image collections,''
  \url{https://www.cs.cornell.edu/~snavely/bundler/}, accessed: 2017-02-28.

\bibitem{openSfM}
J.~E. Solem and Others, ``{OpenSfM: open source structure from motion
  pipeline},'' \url{https://github.com/mapillary/OpenSfM}.

\bibitem{lsdslam}
J.~Engel, T.~{Sch\"ops}, and D.~Cremers, ``{LSD-SLAM: Large-scale} direc
  monocular {SLAM},'' in \emph{European Conference on Computer Vision (ECCV)},
  2014.

\bibitem{dso}
J.~Engel, V.~Koltun, and D.~Cremers, ``Direct sparse odometry,'' in \emph{IEEE
  International Conference on Robotics and Automation (ICRA)}, 2016.

\bibitem{pix4d}
``{Pix4D},'' \url{https://pix4d.com/}.

\bibitem{acute3d}
``{Acute3D},'' \url{https://www.acute3d.com/}.

\bibitem{capturingreality}
``{CapturingReality},'' \url{https://www.capturingreality.com/}.

\bibitem{ham_eccv14}
C.~Ham, S.~Lucey, and S.~Singh, ``Hand-waving away scale,'' in \emph{European
  Conference on Computer Vision (ECCV)}, 2014.

\bibitem{ham}
------, ``Absolute scale estimation of {3D} monocular vision on smart
  devices,'' in \emph{Mobile Cloud Visual Media Computing}.\hskip 1em plus
  0.5em minus 0.4em\relax Springer, 2015, pp. 329--353.

\bibitem{mair}
E.~Mair, M.~Fleps, M.~Suppa, and D.~Burschka, ``Spatio-temporal initialization
  for {IMU} to camera registration,'' in \emph{Robotics and Biomimetics
  (ROBIO), 2011 IEEE International Conference on}.\hskip 1em plus 0.5em minus
  0.4em\relax IEEE, 2011, pp. 557--564.

\bibitem{arun}
K.~S. Arun, T.~S. Huang, and S.~D. Blostein, ``Least-squares fitting of two
  {3-D} point sets,'' \emph{{IEEE} Trans. Pattern Anal. Machine Intell.},
  no.~5, pp. 698--700, 1987.

\bibitem{kanatani}
K.~Kanatani, ``Analysis of {3-D} rotation fitting,'' \emph{{IEEE} Trans.
  Pattern Anal. Machine Intell.}, vol.~16, no.~5, pp. 543--549, 1994.

\bibitem{Kalman:1960}
R.~E. Kalman, ``A new approach to linear filtering and prediction problems,''
  \emph{Transactions of the ASME, Journal of Basic Engineering}, vol.~82,
  no.~1, pp. 35--45, 1960.

\bibitem{Sarkka:2013}
S.~S\"arkk\"a, \emph{Bayesian filtering and smoothing}.\hskip 1em plus 0.5em
  minus 0.4em\relax Cambridge University Press, 2013.

\bibitem{rauch}
H.~E. Rauch, C.~Striebel, and F.~Tung, ``Maximum likelihood estimates of linear
  dynamic systems,'' \emph{AIAA journal}, vol.~3, no.~8, pp. 1445--1450, 1965.

\bibitem{aggarwal}
P.~Aggarwal, Z.~Syed, X.~Niu, and N.~El-Sheimy, ``A standard testing and
  calibration procedure for low cost {MEMS} inertial sensors and units,''
  \emph{Journal of navigation}, vol.~61, no.~02, pp. 323--336, 2008.

\bibitem{shield}
``{NVIDIA Shield tablet},'' \url{https://shield.nvidia.com/tablet/k1},
  accessed: 2016-10-20.

\end{thebibliography}

\end{document}